# Deriving a Minimal *I*-map of a Belief Network Relative to a Target Ordering of its Nodes *

**Izhar Matzkevich**                **Bruce Abramson**
Computer Science Department and Social Science Research Institute
University of Southern California
Los Angeles, CA, 90089-0781
izhar@pollux.usc.edu
bda@pollux.usc.edu

## Abstract

This paper identifies and solves a new optimization problem: *Given a belief network (BN) and a target ordering on its variables, how can we efficiently derive its minimal I-map whose arcs are consistent with the target ordering?* We present three solutions to this problem, all of which lead to directed acyclic graphs based on the original BN's *recursive basis* relative to the specified ordering (such a DAG is sometimes termed the *boundary DAG* drawn from the given BN relative to the said ordering [5]). Along the way, we also uncover an important general principal about arc reversals: when reordering a BN according to some target ordering, (while attempting to minimize the number of arcs generated), the sequence of arc reversals should follow the topological ordering induced by the original belief network's arcs to as great an extent as possible. These results promise to have a significant impact on the derivation of consensus models, as well as on other algorithms that require the reconfiguration and/or combination of BN's.

## 1    INTRODUCTION AND MOTIVATION

Our interest in reconfiguring belief networks (BN's) subject to a specified target ordering grew out of our investigation of combined, or consensus, BN's [2, 3]. We began to consider the problem of combining BN's, because we, like other expert-systems researchers, felt that it was important to develop a mechanism that would allow us to combine the inputs of multiple experts into a single consensus recommendation. The situation that we envisioned was one in which several experts encode independent BN's across the same set of variables. We would like to fuse these BN's together to form a single "consensus" model before any case-specific data (e.g., observations, test-results) is entered. Now, since a BN—by definition—is composed of both an acyclic digraph (DAG) and set of probability distributions, any such consensus BN must include both a consensus structure and a set of consensus numbers. Of these two, structural consensus is both more fundamental, (the consensus structure *houses* the consensus numbers), and less well understood. Our investigation of combined BN's thus began with an investigation of consensus structures.

*Supported in part by the National Science Foundation under grant SES-9106440.

We quickly discovered that the derivation of a consensus DAG is no great trick; a rather straightforward combination of graph union and arc reversal operations accomplish the task in (low-order) polynomial time. The basic idea behind our **FUSE_DAGS** algorithm was that DAG's should be combined sequentially. A topological sort of the first DAG imposes an order on the BN's variables. All other DAG's are expected to conform to that order; those that don't, must reverse the necessary arcs (and add new arcs, where appropriate). DAG's that conform to the same ordering may then be combined using simple graph union [2].

Although **FUSE_DAGS** solves the problem that it was intended to solve, it does have some drawbacks. Paramount among them is that it provides no guarantees about the sparseness of the consensus DAG that it generates. The desirability of sparse consensus DAG's, of course, arises from their ability to better recognize common assumptions of independence and from the intractability of most BN-related algorithms on dense DAG's. A sparse representation of a dependency model (of the type introduced above) is thus always preferable to a denser representation of the same model. Since the sparsity of the DAG constructed by **FUSE_DAGS** is largely a factor of the number of arcs added during arc reversal operations, and the number of arcs reversed is essentially determined by the imposed ordering, the selection of an appropriate ordering appears to be the key to the construction of a sparse consensus model. Unfortunately, it appears that the only way to find an *optimal* ordering is to consider all $N!$ possible orderings (where $N$ is the number of nodes) [4, 8], the derivation of a consensus DAG with a minimal number of arcs is in fact *NP*-hard [3].

Our investigation of consensus structures thus led us to two key questions:

- How can we determine which of two DAG's derived from the *same* dependency model relative to *different* orderings captures *more* of the original dependency model's independencies?

- Given a DAG-isomorphic dependency model and an ordering on its variables, how can we *efficiently* derive a DAG, consistent with the ordering, that captures only the independencies of the original model, and from which no arc may be removed without introducing a spurious independence (i.e., a minimal *I*-map of it which is consistent with the specified ordering)?

The first question defines a more-general relative of the entailment problem [6] for which we have yet to find a solution. The second problem, on the other hand, is answered by applying the right sequence of arc reversals, and is the topic of this paper.

Consider the DAG constructed relative to a target ordering (from some given BN) by drawing arrows *from* every element in the minimal subset of each node's predecessors (according to the target ordering) that render it conditionally independent *to* the node itself. (This construction generates a DAG based on the *recursive basis*, [1] or *causal input list*, [6] of the original BN relative to the specified target ordering.) This newly created DAG is *precisely* the minimal *I*-map of the original dependency model whose arcs are consistent with the target ordering (i.e, the one we wanted).

This paper presents three methods for guiding the selection of arcs to be reversed by two orderings, an initial one gleaned from the (original) BN's topology, and the required target ordering. These methods not only transform the BN into one consistent with the target ordering, but into one that is a minimal *I*-map of the original BN, as well. This is guaranteed, because the resultant DAG is exactly the boundary DAG induced by the recursive basis drawn from the original BN relative to the target ordering. Thus, it is assured to have a minimal number of arcs. Furthermore, it is assured to capture a maximal number of independencies of the original model, while at the same time introducing no spurious ones.

To summarize our underlying motivation, then, the automatic derivation of minimal *I*-maps from a BN

relative to different orderings will help us generate efficient consensus structures.

## 2   ILLUSTRATIONS

Some of the concepts introduced in the previous section may remain somewhat unclear. Before we proceed with the actual algorithms, then, it might be useful to illustrate some of these points with concrete examples.

First, in order to fully understand the impact of node ordering on sparsity, consider the following example. Let evidence variables $b_1, b_2, \ldots, b_n$ be individually dependent on the (unobservable) event $c$. If the nodes are ordered $\alpha_1 = \{c, b_1, \ldots, b_n\}$, the model constructed is DAG $D_{\alpha_1} = (V, \vec{E_1})$, $V = \{c, b_1, \ldots, b_n\}$, $\vec{E_1} = \{(c, b_1), \ldots, (c, b_n)\}$ as shown in Figure 1. $D_{\alpha_1}$ is not only a valid BN, but also a perfect map of the underlying dependency model. Note further that the number of arcs that it contains is *linear* in the number of nodes, ($|\vec{E_1}| = O(|V|)$). The ordering $\alpha_1$ is thus a very good choice. Not all orderings are equal, however. Were the nodes arranged $\alpha_2 = \{b_1, \ldots, b_n, c\}$ instead, the model constructed would have been DAG $D_{\alpha_2} = (V, \vec{E_2})$, $V = \{c, b_1, \ldots, b_n\}$, $\vec{E_2} = \{(b_1, c), \ldots, (b_n, c)\} \cup \{(b_i, b_j)|1 < i < n, i < j \leq n\}$, as shown in Figure 1. $D_{\alpha_2}$, like $D_{\alpha_1}$, is a valid BN and a minimal *I*-map. It is also, however, fully connected; $|\vec{E_2}| = O(|V|^2)$. Furthermore, since $D_{\alpha_2}$ is fully connected, it carries *no* information about conditional independencies. In this instance, then, $D_{\alpha_1}$ is obviously preferable to $D_{\alpha_2}$ (and thus $\alpha_1$ to $\alpha_2$).

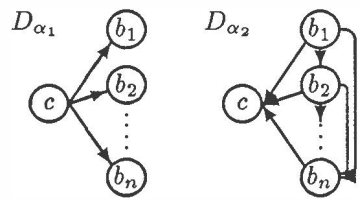

Figure 1: DAGs constructed relative to different orderings from the same dependency model need not be equal.

Second, consider how different sequences of arc reversals may end-up with different *I*-maps of a DAG $D$ relative to the same target ordering $\alpha$. In Figure 2, the input DAG is $D$ and the target ordering $\alpha$ is $\{z, w, y, x\}$. To transform $D$ into a DAG whose arcs are consistent with $\alpha$, one may begin by reversing the arc $(x, y)$. Alternatively, the arc $(x, w)$ may be the first to reverse. The reversal of the arc $(x, w)$ before $(x, y)$ results in a creation of an extra arc $(z, w)$ in $D_2$. The introduction of this new arc could have been avoided if $(x, w)$ were reversed first. $D_1$ is a minimal I-map of $D$ relative to $\alpha$. $D_2$ is not. This kind of "optimal" sequence of reversals is guaranteed by the methods presented in this paper.

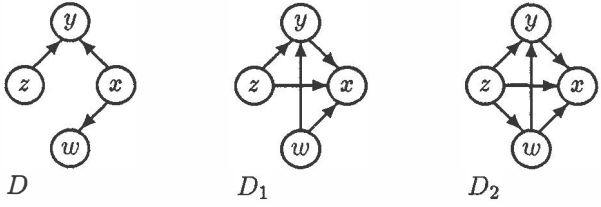

$D$        $D_1$        $D_2$

Figure 2: DAGs constructed relative the same target orderings—yet using different sequences of arc reversals—need not be equal.

## 3  PRELIMINARIES

The basic groundwork upon which our results are based was laid by Pearl and his students in their development of the theory of BN's. The definitions and results presented in this section are taken (albeit with some minor modifications) from their work [4, 5, 8, 6, 1].

A *dependency model* $M$ may be defined over a finite set of objects $U$ as any subset of triplets $(X, Y, Z)$ where $X, Y$ and $Z$ are three disjoint subsets of $U$. $M$ may be thought of as a truth assignment rule for the independence predicate, $I(X, Z, Y)$, read "$X$ is independent of $Y$, given $Z$" (an I-statement of this kind is an *independency*, and its negation a *dependency*). An I-map of a dependency model $M$ is any dependency model $M'$ such that $M' \subseteq M$. A perfect map of a dependency model $M$ is any dependency model $M'$ such that $M' \subseteq M$ and $M \subseteq M'$.

**Definition 1** : *A graphoid is a dependency model that is closed under the following axioms:*
*(i)* **Symmetry** $I(X, Z, Y) \Leftrightarrow I(Y, Z, X).$
*(ii)* **Decomposition** $I(X, Z, Y \cup W) \Rightarrow I(X, Z, Y).$
*(iii)* **Weak union** $I(X, Z, Y \cup W) \Rightarrow I(X, Z \cup W, Y).$
*(iv)* **Contraction** $I(X, Z, Y)$ & $I(X, Z \cup Y, W) \Rightarrow I(X, Z, Y \cup W).$
*A graphoid is intersectional if it also obeys the following axiom:*
*(v)* **Intersection** $I(X, Z \cup Y, W)$ & $I(X, Z \cup W, Y) \Rightarrow I(X, Z, Y \cup W).$

Examples of graphoids include the *probabilistic dependency models* and the *acyclic digraph (DAG) models*. The criterion necessary for a DAG to capture an independence model is known as d-separation.

**Definition 2** *[6]: If $X$, $Y$, and $Z$ are three disjoint subsets of nodes in a DAG $D$, then $Z$ is said to d-separate $X$ from $Y$, denoted $I(X, Z, Y)_D$, if and* only if there is no path [1] *from a node in $X$ to a node in $Y$ along which the following two conditions hold: (i) every node with converging arrows either is or has a descendant in $Z$, and (ii) every other node is outside $Z$. A path satisfying the conditions above is* **active**, *otherwise it is* **blocked** *by $Z$.*

Whenever a statement $I(X, Z, Y)_D$ holds in a DAG $D$, the predicate $I(X, Z, Y)$ is said to be graphically verified (or an independency), otherwise it is graphically unverified by $D$ (or a dependency). The d-separation criterion is a powerful qualitative method for identifying conditional independencies in BN's. By applying the d-separation criterion on a BN's underlying DAG, one is able to identify all (and only) independencies induced by this BN.

Given some dependency model $M$ on a finite set of variables $V$, let $\alpha$ be some complete ordering on $V$. On $M$, define a set $L_\alpha = \{I(x, B(x), U(x) \setminus B(x)) | x \in V\}$. If $U(x)$ is the set of variables preceding $x$ in the ordering $\alpha$, then $B(x) \subseteq U(x)$ would be a minimal subset of them rendering $x$ conditionally independent from the rest of $U(x)$.

The set $L_\alpha$ is termed the *recursive basis* drawn from $M$ relative to $\alpha$ [1]. If $M$ is an intersectional graphoid (e.g., a DAG), $L_\alpha$ is unique.

Given a set $L_\alpha$, one can generate a DAG by pointing an arc from each $y \in B(x)$ to $x$ for each $x \in V$. If we denote $L_\alpha$'s closure under the (intersectional) graphoid axioms $CL(L_\alpha)$ and the DAG generated $D_\alpha$, we get the following theorem:

**Theorem 1** *[8]: $D_\alpha$ (with the d-separation criterion) is a perfect map of $CL(L_\alpha)$.*

Furthermore, if we attempt to capture all independencies in a graphoid $M$, we can do it by considering the collection of the $D_\alpha$'s taken over all possible orderings $\alpha$ on $V$:

**Theorem 2** *[8]: If a dependency model $M$ is a graphoid, then the set of DAGs generated from all recursive bases of $M$ is a perfect map of $M$ if the criterion for separation is that d-separation must exist in one of the DAGs.*

Assume we are given two DAGs $D_1 = (V, \vec{E_1})$ and $D_2 = (V, \vec{E_2})$. In order for us to determine if one *entails* the other, namely, if one is an I-map of the other, Pearl and his students provide us with the following criterion [6]:

**Theorem 3** : *A necessary and sufficient condition for a diagram $D_1$ to entail $D_2$ is that, for every node $x_i$ having parents $B_2(x_i)$ in $D_2$ we have* $I(x_i, B_2(x_i), U_2(x_i) \setminus B_2(x_i)) \in D_1$, *where* $U_2(x_i) = \{x_1, \ldots, x_{i-1}\}$ *is a set of predecessors of* $x_i$ *in some ordering that is consistent with the arrows of* $D_2$.

---
[1] By path we mean a sequence of edges in the underlying undirected graph, i.e., ignoring the directionality of the arcs.

Simply put, if all the independencies contained in a recursive basis of $D_2$ drawn relative to some complete ordering which is consistent with $\vec{E_2}$ are verified graphically in $D_1$, we need look no further; we are assured that all other independencies revealed in $D_2$ are in $D_1$ as well.

## 4    DERIVING DAG'S

A closely related problem encountered during our work is the following: "given a DAG $D = (V, \vec{E})$ (or any DAG-isomorphic dependency model), and some complete ordering $\alpha$ on $V$, how to can we efficiently derive a DAG who is a minimal DAG of $D$ relative to $\alpha$?". In simple words, given $D$ and $\alpha$, how can we derive the minimal DAG (over $V$) whose arcs are consistent with $\alpha$, for which each graphically verified independency is graphically verified in $D$ as well. This DAG is exactly the one induced by the relevant recursive basis drawn from $D$ relative to $\alpha$.

An efficient solution to this problem is highly important because it may be used in a procedure to derive sparse consensus models. The DAG constructed from the recursive basis drawn from the input DAG $D$ relative to $\alpha$ is a minimal $I$-map of $D$. No arc can be removed without destroying the $I$-mapness property; it is therefore assured to have a minimal number of arcs among all $I$-maps of $D$ which are consistent with the ordering $\alpha$. It is also assured to capture the maximal number of independencies among them.

Given a DAG $D = (V, \vec{E})$ and a complete ordering $\alpha$ on $V$, **METHOD A** (below) may be used to derive the relevant recursive basis (and the DAG $D_\alpha$). To begin, some preliminary work is required on $D$. First, we construct two sequences over $V$'s elements. One of the sequences (termed $S_\alpha$) is composed of $V$'s elements ordered according to $\alpha$ in a decreasing order left-to-right. The other sequence (termed $S_\beta$) is constructed by iteratively "reducing" $D$ (or a copy of it thereof) via an ordered elimination of its nodes (and all arcs pointing into them) *one at a time*. Let $Q_D(x)$ denote the set of immediate successors of a node $x$ in a DAG $D$, then $S_\beta$ is constructed using the following procedure:

### CONSTRUCTING $S_\beta$

1. Initiate $S_\beta = [\ ]$, $D' = (V' \leftarrow V, \vec{E'} \leftarrow \vec{E})$.
2. **do until** $V'$ is exhausted
   Select an element $y \in V'$ for which $Q_{D'}(y) = \emptyset$ (i.e., $y$ is a *barren node* in $D'$), and $y \notin S_\beta$.
   Percolate $y$ in $S_\beta$ **left-to-right**.
   **do for each** $z \in S_\beta$ encountered

If $z \in Q_D(y)$ **stop**;
/* note: $z$ is checked in $D$, not in $D'$ */
else, if $z$ is to the right of $y$ in $S_\alpha$ **stop**.
If stopped, "plug" $y$ immediately to the left of $z$ in $S_\beta$, and exit the inner **do** loop.
/* i.e., $S_\beta = [\ldots yz \ldots]$ */
**enddo**
If $S_\beta$ is exhausted, **stop** and "plug" $y$ as $S_\beta$'s new right-most element.
$V' \leftarrow V' \setminus \{y\}$, $\vec{E'} \leftarrow \vec{E'} \setminus \{(z, y) | z \in V'\}$.
**enddo**

The sequence $S_\beta$ constructed is consistent left-to-right with the partial (topological) ordering initially induced by $\vec{E}$ on $V$. Furthermore, as is explained below, it is "closer" to $S_\alpha$ than any other sequence which is consistent with that partial ordering.

The notion of *percolation* deserves some explanation. Percolating an element left-to-right in a sequence is done by repeatedly interchanging it with its immediate neighbor to the right. An *interchange operation* over two elements $x$ and $y$ in a sequence $S$ is a one by which $x$ and $y$ replace each other in $S$. If $S = [\ldots x \ldots y \ldots]$ before the interchange, then $S = [\ldots y \ldots x \ldots]$ after the interchange; no other element in $S$ is involved. Given an input DAG $D = (V, \vec{E})$ and an ordering $\alpha$ on $V$, the method for deriving the DAG $D_\alpha$ induced by the recursive basis drawn relative to $\alpha$ proceeds as follows:

### METHOD A

1. Given $D = (V, \vec{E})$ and an ordering $\alpha$, form the two sequences $S_\alpha$, $S_\beta$.
2. Initialize $S = S_\beta$.
3. **do until** $S = S_\alpha$
   Find the **left-most** element in $S$ which should interchange with its adjacent **left** neighbor and percolate it right-to-left as much as possible (via interchange operations) relative to $S_\alpha$. Let $x$ the element propagated, and $xy$ the pair interchanged in $S$. If $(y, x) \in \vec{E}$, reverse it in $D$ (perform an arc reversal operation).
   **enddo**

In fact, (3.) may further be simplified:

3. **do until** $S = S_\alpha$
   Interchange the **left-most** pair of adjacent elements $\ldots xy \ldots$ in $S$ for which $y$ appears somewhere to the **left** of $x$ in $S_\alpha$.
   If $(x, y) \in \vec{E}$, reverse it in $D$ (perform an arc reversal operation).
   **enddo**

In any case, the resultant DAG obtained through applying the sequence of arc reversals on $D$ is exactly the desired $D_\alpha$. Consider again the example of Fig-ure 2. Recall that the input DAG was $D$ and the target ordering $\alpha$ was $\{z, w, y, x\}$. Two possible initial orderings are $\alpha_1 = \{z, x, w, y\}$ and $\alpha_2 = \{z, x, y, w\}$. If ordering $\alpha_2$ is selected, $x$ and $y$ would be interchanged first, and the resultant DAG would be $D_2$. If, on the other hand, $\alpha_1$ is the initial ordering selected (as guaranteed by **METHOD A)**, $D_1$ would be the resultant DAG. $D_1$ is a minimal *I*-map of $D$ relative to $\alpha$. $D_2$ is not.

Note that in the discussion above, we related $S_\beta$ and $S_\alpha$ by the notion of "closer than any other relevant sequence". This relates to the following property: no "unnecessary" interchange operation is ever required on two adjacent elements $xy$ in $S$. A pair of adjacent elements in $S$ is interchanged if and only if $x$ is a real descendant of $y$ in $D$, yet $x$ is to the left of $y$ in $S_\alpha$. This property translates into the following fact: no unnecessary arc reversal is ever required on $D$. An unnecessary arc reversal is found in Figure 2. The reversal of $(y, w)$ in $D_2$ was unnecessary. Since an arc reversal may involve the creation of new arcs (thereby the elimination of independencies), redundant reversals are unwanted.

Our aim is to prove that the method is correct, or in other words, that: (*i*) the algorithm halts, (*ii*) at any arc-reversal the digraph $D$ "rearranged" is acyclic, (*iii*) the resulted DAG is consistent with the target ordering $\alpha$, and (*iv*) the resulted DAG is the same as the one induced by the recursive basis drawn from $D$ relative to $\alpha$.

In proving all that, we may also get the following: (*i*) given a target ordering $\alpha$ and a DAG $D$, the algorithm **FUSE_DAGS** presented in [2] may be utilized to derive a DAG which is the same as the one induced by the recursive basis drawn from $D$ relative to $\alpha$, and (*ii*) the same holds if we utilize the "bubble-sort-like" interchange algorithm presented in [7] (that method is termed below **METHOD B**).

Let us then begin sketching the proof. Since the number of interchange operations is bounded by $|V| \cdot (|V| - 1)$ (a pair of elements can only be interchanged once), the algorithm obviously halts. Furthermore, at any time all arcs in $D$ point "rightward" in $S$ (i.e., no arc can point from a node $x$ to a node $y$ if $x$ is to the **right** of $y$ in $S$). This property guarantees that the sequence of arc reversals on $D$ (induced by the order by which elements in $S$ are interchanged) preserves its acyclicity at any point during. Finally, Shachter [7] showed that we can get from any initial ordering to any target ordering through a sequence of interchange operations among adjacent elements in a sequence. All these combine to prove that the DAG constructed is one for which the partial ordering that its arcs induce on its nodes is consistent with $\alpha$. In the next section we explain why the end-result of the method is indeed the DAG induced by the relevant recursive basis.

## 5  PROOF OF CORRECTNESS

We begin with some preliminary notation. Let $D_\alpha$ be the DAG induced by the recursive basis drawn from the input DAG $D$ relative to $\alpha$. Let $D_m$ be the DAG as it stands after $m$ interchange operations according to the method (**A**) presented. To construct a proof, suffice it to show that $D_\alpha$ is an *I*-map of $D_m$ for all $m \geq 0$ (it is in fact a minimal such an *I*-map). This is enough for if $D_{\bar{m}}$ is the DAG finally obtained by our method, $D_{\bar{m}}$ is consistent with $\alpha$. $D_\alpha$ is an *I*-map of $D_{\bar{m}}$ which in turn is an *I*-map of the input DAG $D$ (the *I*-mapness property is preserved through each interchange operation among neighbors). Finally, $D_\alpha$ is a *minimal I*-map of $D$. These statements combine (via the "sandwich rule") to prove that $D_{\bar{m}} = D_\alpha$.

The proof requires some simple claims. For the sake of clarity, most of them are omitted. Two of them are:

**Claim 1** : *Given sets of nodes* $\{x\}, Y, Z, S_1, S_2 \subseteq V$, *if (i)* $x$ *and* $Y$ *not d-separated given* $S_1$, *and (ii)* $x$ *and* $Z$ *not d-separated given* $S_2$, *then:*
1. *If* $x$ *is a* **head-to-head** *node relative to some trail* [2] *from* $x$ *to* $Y$ *activated by* $S_1$ *and some trail from* $x$ *to* $Z$ *activated by* $S_2$, *then* $Y$ *and* $Z$ *are* **not** *d-separated given* $\{x\} \cup S_1 \cup S_2$. *If no such trails exist for which* $x$ *is not a head-to-head node, then* $Y$ *and* $Z$ *are* **not** *d-separated by* $S_1 \cup S_2$.

In fact, Claim 1 is a restatement of the *weak-transitivity* axiom (or actually, its contrapositive) [5]:

$$I(X, Z, Y) \text{ \& } I(X, Z \cup \{\gamma\}, Y) \Rightarrow I(X, Z, \{\gamma\}) \vee I(\{\gamma\}, Z, Y)$$

**Claim 2** : *Given two DAGs* $D_1$ *and* $D_2$, *if* $D_1$ *is an I-map of* $D_2$ *(relative to the d-separation criterion), then given sets of nodes* $\{x\}, Y$ *and* $S$, *if* $\{x\}$ *and* $Y$ *are* **not** *d-separated given* $S$ *in* $D_2$, *the same holds in* $D_1$.

Claim 2 is almost definitional. These claims lead directly to the following theorem:

**Theorem 4** : *Let* $D_\alpha$ *be the DAG induced by the recursive basis drawn from* $D$ *relative to* $\alpha$. *Let* $D_{\bar{m}}$ *be the DAG constructed by the method presented above given* $D$ *and* $\alpha$. *Then* $D_\alpha = D_{\bar{m}}$.

**Sketch of a proof:** The proof uses induction on $m$, the number of interchange operations required to transform $S_\beta$ (the initial ordering) into $S_\alpha$ (the target ordering). If $m = 0$, the case is obvious. So is the case for $m = 1$; an assumption to the contrary leads to an immediate contradiction through Claim 1. Let us assume that the proposition holds after $m \geq 1$ interchange operations (i.e., $D_\alpha$ is an $I$-map of $\tilde{D}_m$). Consider $D_{m+1}$, the DAG obtained from $D_m$ following the $m + 1$ interchange operation. We show that $D_\alpha$ is an $I$-map of $D_{m+1}$. Let $xy$ denote the two variables interchanged in the $m + 1$ step. If this interchange operation does not require an actual reversal of $(x, y)$ in $D$, then $D_{m+1} = D_m$, and we are done.

---

[2] By trail we mean a sequence of links that form a path in the underlying undirected graph.

Assume that arc $(x, y)$ is reversed. For $D_\alpha$ not to be an $I$-map of $D_{m+1}$ (Claim 2) there must exist three sets of variables $X, Y, Z \subseteq V$ for which $X, Y$ are $d$-separated given $Z$ in $D_\alpha$, yet not in $D_{m+1}$. Since $D_\alpha$ is an $I$-map of $D_m$, $X$ and $Y$ are $d$-separated given $Z$ in $D_m$ as well. (With the notation given above, we therefore have $I(X, Z, Y)_{D_\alpha}$, $I(X, Z, Y)_{D_m}$, and yet $\neg I(X, Z, Y)_{D_{m+1}}$.) For us to consider trails in the relevant DAGs, suffice it to consider only *simple trails*—trails that form no cycles in the underlying undirected graph [1]. We would, in fact, be interested in identifying those cases in which (i) no trail between any two nodes $\tilde{x} \in X$, $\tilde{y} \in Y$ is active given $Z$ in $D_m$, (ii) a trail between $\tilde{x} \in X$ and $\tilde{y} \in Y$ is activated in $D_{m+1}$ given $Z$, and yet (iii) no trail among these two nodes is active given $Z$ in $D_\alpha$.

Three relevant generic cases are illustrated in Figure 3:

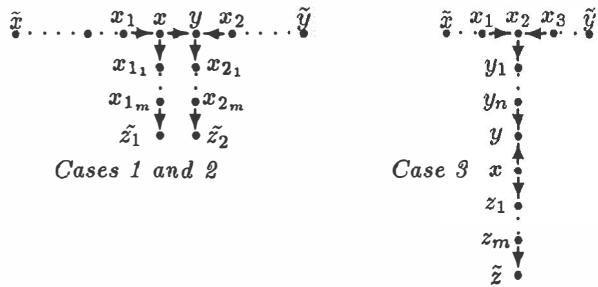

Cases 1 and 2          Case 3

Figure 3: The three relevant cases.

1. In the first case, assume $x, \tilde{z}_2 \in Z$ (it is possible that $\tilde{z}_2 = y$, i.e., when the chain $y \to \ldots \tilde{z}_2$ is of length 0). The reversal of $(x, y)$ renders the trail $\tilde{x} \cdots x_1 \to y \leftarrow x_2 \cdots \tilde{y}$ active. We assume no trail between $\tilde{x}$ and $\tilde{y}$ is active given $Z$ in $D_\alpha$.

2. In the second case, assume $\tilde{z}_1 \in Z$ (it is possible that $\tilde{z}_1 = x$, i.e., when the chain $x \to \ldots \tilde{z}_1$ is of length 0), and $y \notin Z$. The reversal of $(x, y)$ renders the trail $\tilde{x} \cdots x_1 \to x \leftarrow x_2 \cdots \tilde{y}$ active. We assume no trail between $\tilde{x}$ and $\tilde{y}$ is active given $Z$ in $D_\alpha$.

3. In the third case, assume $\tilde{z} \in Z$ (and it is possible that $y = x_2$ or $x = \tilde{z}$, i.e., when any one of the chains $x_2 \to \ldots y$ or $x \to \ldots \tilde{z}$ is of length 0). The reversal of $(x, y)$, renders the trail $\tilde{x} \cdots x_1 \to x_2 \leftarrow x_3 \cdots \tilde{y}$ active. We again assume no trail between $\tilde{x}$ and $\tilde{y}$ is active given $Z$ in $D_\alpha$.

In order to prove that $D_\alpha$ is an $I$-map of $D_{m+1}$, suffice it to show that in any of these three cases, $\tilde{x}$ and $\tilde{y}$ can not be $d$-separated by $Z$ in $D_\alpha$ (i.e., there exists some trail between them which is activated by $Z$ in $D_\alpha$), thus establishing a contradiction. This, or else the induction hypothesis ($D_\alpha$ being an $I$-map of $D_m$), Theorem 3, and/or one of the two Claims given above, are violated. A proof for any of these cases is somewhat tedious, and precluded by space limitations.

To summarize the discussion to this point, $D_\alpha$ is an $I$-map of $D_m$, the DAG obtained after $m$ interchange operations ($m \geq 1$). By the line of reasoning given above, we therefore get that $D_\alpha = D_{\bar{m}}$, the DAG constructed by applying the method (**A**) on $D$, as required.

Theorem 1 also has two corollaries. The first concerns the algorithm **FUSE_DAGS** presented in [2]; it requires a change in the conditions by which we select the next arc to reverse so that an actual *transitive closure* is considered every time the algorithm requires only *topological values*. Now, given a DAG $D_1 = (V, \vec{E})$, and some complete ordering $\alpha$ on $V$, construct a DAG $D_2$ by generating a **directed chain** (i.e., a linear descendant list) on $V$'s elements according to $\alpha$. If $D_1$ and $D_2$ are given as input for the algorithm **FUSE_DAGS**, let the output DAG be $(V, \vec{E}_1' \cup \vec{E}_2)$, where $\vec{E}_1'$ is the set of arcs in $D_1$ after it is "rearranged" via a sequence of arc reversals according to the topological ordering induced by $\vec{E}_2$ on $V$. We then claim:

**Corollary 1** : $D_\alpha = (V, \vec{E}_1')$.

Corollary 1 is proven by examining the conditions defined in **FUSE_DAGS** by which the next candidate for reversal is selected. It can be shown that they determine a sequence of arc-reversals which would be obtained if our method of "always interchange the left-most candidates" above is applied, starting with a sequence $S_\beta$ constructed according to the procedure given above.

Another algorithm (rather similar to **METHOD A**) is obtained by utilizing the "bubble-sort-like" interchange algorithm presented in [7]. Given $D = (V, \vec{E})$ and some target ordering $\alpha$, we begin by constructing the sequences $S_\alpha$ and $S_\beta$ as above. The algorithm itself is:

**METHOD B**

1. Given $D = (V, \vec{E})$ and an ordering $\alpha$, form the two sequences $S_\alpha$, $S_\beta$.
2. Initialize $S = S_\beta$.
3. do until $S = S_\alpha$
   Find the **left-most** element in $S$ which should interchange with its adjacent **right** neighbor and percolate it left-to-right as much as possible (via interchange operations) relative to $S_\alpha$. Let $x$ be the element propagated, and $yx$ the pair interchanged in $S$. If $(x, y) \in \vec{E}$, reverse it in $D$ (perform an arc reversal operation).

**enddo**

For this "bubble-sort-like" interchange algorithm, if $D_m$ is the resulted DAG, then:

**Corollary 2** : $D_\alpha = D_m$.

A proof is obtained through the following simple claim:

**Claim 3** : *If $xy$ is a pair of adjacent neighbors in $S$ to be interchanged when* **METHOD B** *is applied, and if $(x, y)$ is to be reversed in $D$ as a result, then:*

1. *$(x, y)$ would have been reversed if* **METHOD A** *was applied as well (with the same $S_\beta$ and $S_\alpha$ sequences), and*

2. *The sets of immediate predecessors of $x$ and $y$ in $D$ are the same at the time $(x, y)$ is reversed, whether* **METHOD A** *or* **METHOD B** *were applied (with the same $S_\beta$ and $S_\alpha$ sequences again).*

Alternatively, a proof similar to the one given for **METHOD A** above may again be constructed by a direct induction on the number of interchange operations required to transform $S_\beta$ to $S_\alpha$.

A quick complexity assessment is now required. A more thorough assessment is found in [2]. The evaluation of $S_\beta$ given a DAG $D$ and a target ordering $\alpha$ requires only a topological ordering of $D$. This ordering is obtained by a topological sort in $O(\vec{E})$ steps. Since each interchange operation may be followed by an actual reversal of an arc in $D$, there are $O(|V|^2)$ potential arc reversals. Since an arc between any pair of nodes $x, y \in V$ may only be created (or reversed) at most once, the overall complexity of the methods given above is therefore $O(|V|^2)$.

# 6   SUMMARY

In this paper we identified an optimization problem that grew out of our interest in sparse consensus models:

- Given a DAG-isomorphic dependency model (i.e., a BN) and a target ordering, how can we efficiently derive a minimal I-map of it whose arcs are consistent with the target ordering?

We presented three solutions to this problem, all of which may prove to be useful in our construction of sparse consensus models. We also discovered an important general principal about arc reversals:

- When reordering a BN according to some target ordering, for the purpose of minimizing the number of arcs generated, the sequence of arc reversals should follow the topological ordering induced by the original BN's arcs to as great an extent as possible.

All three methods generate the minimal I-map of the original BN whose arcs are consistent with the target ordering. The derived DAG, in fact, is exactly the *boundary DAG* induced by the recursive basis drawn from the original BN relative to the said target ordering. The DAG formed from the recursive basis thus has the desired properties: it maximizes the number of the (original) independencies captured, and it has the minimal number of arcs among all other I-maps of the original BN whose arcs are consistent with the given target ordering.

# Acknowledgement

We thank Ross Shachter for the insight and a useful feedback on an earlier draft of this paper.

# References

[1] D. Geiger, T. Verma, and J. Pearl. *Identifying independence in Bayesian networks.* In *Networks*, Vol. 20:507–534, 1990.

[2] I. Matzkevich and B. Abramson. The topological fusion of Bayes nets. In *Proceedings of the Eighth Conference on Uncertainty in Artificial Intelligence*, 191–198, July 1992.

[3] I. Matzkevich and B. Abramson. Some complexity considerations in the combination of belief networks. In this volume, 1993.

[4] J. Pearl and T. Verma. The logic of representing dependencies by directed graphs. In *Proceedings of the Sixth National Conference on Artificial Intelligence*, 374–379, July 1987.

[5] J. Pearl. *Probabilistic reasoning in intelligent systems: networks of plausible inference.* Morgan Kaufmann, San Mateo, CA, 1988.

[6] J. Pearl, D. Geiger, and T. Verma. *The logic of influence diagrams.* In R. M. Oliver and J. D. Smith (editors), *Influence Diagrams, Belief Networks and Decision Analysis*, 67–87, John Wiley and Sons, Ltd., Sussex, England 1989.

[7] R. D. Shachter. *An ordered examination of influence diagrams.* In *Networks* 20:535–564, 1990.

[8] T. Verma and J. Pearl. Causal networks: semantics and expressiveness. In *Proceedings of the Fourth Workshop on Uncertainty in Artificial Intelligence*. 352–359, St. Paul, Minnesota, 1988.